\documentclass[journal]{IEEEtran}
\usepackage{amsmath,amsfonts}
\usepackage{algorithmic}
\usepackage{algorithm}
\usepackage{array}
\usepackage[caption=false,font=normalsize,labelfont=sf,textfont=sf]{subfig}
\usepackage{textcomp}
\usepackage{stfloats}
\usepackage{url}
\usepackage{verbatim}
\usepackage{graphicx}
\usepackage{cite}
\usepackage{amsthm}
\usepackage{amsmath}
\usepackage{amssymb}
\usepackage{txfonts}
\usepackage{multirow}

\usepackage{bbm}
\usepackage{bm}

\begin{document}

\title{Looking From the Future: Multi-order Iterations Can Enhance Adversarial Attack Transferability}

\author{Zijian Ying, Qianmu Li, Tao Wang, Zhichao Lian, Shunmei Meng, Xuyun Zhang
\thanks{Zijian~Ying, Zhichao~Lian, Qianmu~Li are with the School of Cyber Science and Technology, Nanjing University of Science and Technology, Nanjing 210094, China. E-mail: zjying@njust.edu.cn, lzcts@163.com, qianmu@njust.edu.cn.}
\thanks{Tao~Wang and Meng~Shunmei are with the School of Computer Science and Technology, Nanjing University of Science and Technology, Nanjing 210094, China. E-mail: 122106010829@njust.edu.cn, mengshunmei@njust.edu.cn}
\thanks{Zhang~Xuyun is with the School of Computing, Faculty of Science and Engineering, Macquarie University, NSW 2109, Australia. E-mail:
 xuyun.zhang@mq.edu.au}
}
\maketitle

\begin{abstract}
Various methods try to enhance adversarial transferability by improving the generalization from different perspectives. In this paper, we rethink the optimization process and propose a novel sequence optimization concept, which is named Looking From the Future (LFF). LFF makes use of the original optimization process to refine the very first local optimization choice. Adapting the LFF concept to the adversarial attack task, we further propose an LFF attack as well as an MLFF attack with better generalization ability. Furthermore, guiding with the LFF concept, we propose an $LLF^{\mathcal{N}}$ attack which entends the LFF attack to a multi-order attack, further enhancing the transfer attack ability. All our proposed methods can be directly applied to the iteration-based attack methods. We evaluate our proposed method on ImageNet1k datasets by applying several SOTA adversarial attack methods under four kinds of tasks. Experimental results show that our proposed method can greatly enhance the attack transferability. Ablation experiments are also applied to verify the effectiveness of each component. The source code will be released after this paper is accepted.
\end{abstract}

\begin{IEEEkeywords}
Adversarial attack, Optimization process, XAI
\end{IEEEkeywords}

\section{Introduction}
Deep Neural Networks(DNNs) have shown vulnerability to adversarial examples. 
By adding human-imperceptible perturbations to the clean input, DNNs will result in misclassification. 
At the same time, some adversarial examples generated on one network can also have an attack effect on another network. 
This phenomenon is called adversarial attack transferability. 
And, some methods are trying to attack a black-box network by generating adversarial examples on a white-box network. This kind of attack behavior is called the transfer attack.
For the generated adversarial sample, the more black-box networks it can successfully attack, the stronger the transfer attack capability of the adversarial sample will be.
This capability is also called adversarial transferability.

Nowadays, a lot of works try to enhance adversarial transferability with various kinds of approaches. One of the most classic methods is the MI-FGSM\cite{dong2018boosting}. MI-FGSM, utilizing momentum to replace the gradient, inherits the iterative method in I-FGSM\cite{kurakin2018adversarial} to enhance the attack capability of the adversarial examples and enhance adversarial transferability by combining historical gradient information at the same time. This leads to a branch of adversarial attack methods with optimizing gradients themselves. These methods attempt to improve the adversarial transferability of adversarial examples by changing the gradient of the optimization process to avoid falling into local minima. Then another branch of adversarial methods is directly modifying the loss function to change the optimization terminal. Recently, input-transformation methods utilize data augmentation approaches, changing the input, to improve the diversity of the gradients. These methods indeed effectively enhance the adversarial transferability.

However the essential of these methods is solving the optimization problem, i.e. optimization methods. 
The optimization process of finding the optimal result through iteration can be regarded as a sequence.
Then, each search point is a node in this sequence.
Once the optimization method and the starting point of the search are determined, the sequence itself is relatively fixed.
It can be found that most of the existing adversarial attack methods only generate the next node with information from the current node and previous nodes. 
This will prevent the optimization process itself from benefiting from information from subsequent nodes.
Merely using the current node or previous nodes will make each node update only in the direction of local optimization with limited global information.
Since the subsequent nodes belong to the future for the current node, we name the information from the subsequent nodes as future information.
Similarly, the information from previous nodes is called historical information.
Therefore, refining the optimization process with future information is necessary.

There are a few works that have made preliminary attempts to use this future information. \cite{wang2021boosting} proposed the PI-FGSM method which looks ahead with one step to guide the momentum generation. Although this work somehow utilizes a little future information, i.e. the gradient from the next node, the terminal goal is to rich the local information by sampling around the current node. Another work that somehow utilizes future information is \cite{wang2022boosting}.
This work determines the starting state of momentum to optimize gradient consistency by pre-querying the perturbations after N rounds before starting the search. 
However, the future information is merely utilized once at the very beginning of the optimization. Also, with the search steps going greater, the future information will be diluted.
Therefore, the above two works still only use local gradients to update nodes to a certain extent.

In this work, we start by drawing the concept of looking from the future (LFF). LFF rethinks the optimization process and regenerates the first node from the perspective of the optimization result. 
However, there are two problems for LFF directly applying to generate adversarial examples. 
The first one is the constraint of perturbation of each node makes the re-optimization result from LFF have worse attack capability.
Another one is if the optimization process utilizes the historical information, LFF will repeatedly accumulate historical information, thus falling into local search again.
To tackle the above two problems, we modify the LFF concept into the process of the original adversarial attack methods and formally propose the LFF attack.
We reintegrate the weights of each piece of information in the optimization process so that no overfitting of local optimization will occur regardless of whether only the current node information or historical information is used.
Therefore, the LFF attack can be easily applied to the existing adversarial attack methods, which are based on the iteration optimization process, to enhance the adversarial attack performance.
At the same time, to reduce the complexity of the algorithm and further increase the adversarial transferability, we also combined the existing mechanisms to optimize the attack process. Furthermore, we also propose a multi-order LFF attack, under the guidance of the LFF concept, which further enhances the transfer attack performance.

To evaluate our method comprehensively, we evaluate the performance of our proposed method with several state-of-the-art adversarial attack methods on a wide range of high-performance deep neural networks, containing both CNN structure networks and Transformer structure networks. By covering the scenery of single model attacks, attack defensive networks, targetted attacks and ensemble network attacks, the comparison experiment can verify the effectiveness of our proposed method. The ablation experiment is also applied to check the effect of each component and each hyperparameter.

The remainder of this paper is structured as follows. Section II delves into related work on adversarial attack methods.
Section III provides an overview of the notations used in adversarial attacks, the concept of looking from the future and the difficulty of applying LFF to the adversarial attack process.
In Section IV, we present the LFF attack method and derivation process. The multi-order LFF attack method is also attached.
Section V presents the results of our experiments and corresponding analysis. 
Lastly, in section VI, we offer some concluding remarks and outline potential avenues for future research.

\section{Related Work}
Since \cite{goodfellow2014explaining} proposed the FGSM methods, the adversarial attack, especially the transfer attack, has been greatly improved. 
A lot of methods improve the adversarial attack based on the FGSM methods which are marked as FGSMs. 
One of the most classic methods is the I-FGSM\cite{kurakin2018adversarial}.
I-FGSM enhances the adversarial attack by iteratively searching the perturbation, while FGSM merely generates the adversarial example with one step.
Although I-FGSM can successfully attack the white-box networks, which means a high attack capability, it has a very poor transfer attack performance.
Then various branches of methods try to enhance the adversarial transferability based on the I-FGSM.

\subsection{FGSMs based on gradients }
One of the branches is modifying the gradients to obtain generalization adversarial examples.
The most representative work is the MI-FGSM\cite{dong2018boosting}.
Via introducing momentum into to adversarial examples generation process, gradients from each previous node will be utilized to smooth the search direction for the current node.
This will help the optimization process somehow avoid the local minima.
NI-FGSM\cite{lin2019nesterov} method refined the momentum calculation process to further enhance the adversarial example generalization.
VMI-FGSM and VNI-FGSM \cite{wang2021enhancing} utilize the variance tuning the momentum to further avoid the local minima.
EMI-FGSM\cite{wang2021boosting} sampling the local gradient information around the node from PI-FGSM and ensemble the information to enhance the adversarial transferability.
PGN\cite{ge2023boosting} tries to find the local maxima point to directly find the better transfer attack condition.
MIG\cite{ma2023transferable} utilizes the integrated gradients in XAI methods to enhance the gradient in MI-FGSM.
Some other works inspired by the optimizers refine the gradients or the momentums with the latest optimizer \cite{zhang2023generate, yang2023adversarial,zhang2023boosting}. 

The above methods mainly focus on refining the gradient itself, while some other methods try to enhance the gradient by modifying the loss function.
\cite{Li_2020_CVPR} proposes the Po+Trip loss to enhance the targeted transfer attack inspired by the Poincare distance. 
\cite{zhao2021success} thinks that the simple logit loss also has good transfer attack performance.
\cite{zhang2022investigating} considers the class interaction and proposes the relative Cross Entropy loss by raising the prediction probability of other classes to enhance the adversarial attack generalization.

\subsection{FGSMs based on data augmentation }
Another one of the branches is utilizing data augmentation to enhance the diversity of input, which is also called input transformation.
The most representative work is the $\text{DI} ^ 2$-FGSM \cite{xie2019improving}, which uses resize and padding operations to enhance input diversity.
TI-FGSM \cite{dong2019evading} utilizes shifting operation with a kernel matrix on the gradients. 
SI-FGSM \cite{lin2019nesterov} applies scale transformation to the input image to gain augmented data.
SIA \cite{wang2023structure} divides the original image into blocks and applies a random image transformation onto each image block to craft a set of diverse images for gradient calculation.
BSR \cite{wang2024boosting} furthermore introduces the block shuffle and rotation operations into the input transformation.

The above methods mainly focus on the transformation of the input image itself, which can be called self-transformation. Some other methods try to enhance the input diversity by introducing information from other images. \cite{wang2021admix} extends the SI-FGSM with the mixup strategy, which is named Admix. This method chooses several other images and mixes them into the original images to enhance the input diversity. \cite{wang2023boost} improves the mixup strategy with a non-linear way to mask the chosen image into the original image. \cite{wang2023rethinking} rethinks the process of the admix and changes the mixing images process to the mixing gradients process.

\section{Preliminary}

\subsection{Notations}
Here we first give out some fundamental notations. 
$ x\in R^N $ is the input data sample, where $N$ is the data dimension. 
$ H : R^N \to R^D$ is the feature extractor, where $D$ is the feature dimension. 
$ G : R^D \to R^K$ is the classifier, where $K$ is the number of classes. 
Then $ F = G \circ H: R^N \to R^K $ is the entire classification network. 
The output of $F$ is $ z \in R^K$ called logits. 
The output of $\text{softmax}(z)$ is $p$ which is the predicted probability. 
$p$ is also marked as $P_F(x) \in R^K $ to emphsize the network and input data. 
The goal of the adversarial attack is to generate an adversarial example $\hat{x}$ that can lead to $f$ fail, which is expressed as $F(x) = y_t \wedge F(\hat{x}) \neq y_t$. 
Here $y_t$ indicates the truth label of $x$ as the specific predicted class index. 
To express without ambiguity, 
$y$ represents the index classification result for $F$, $z$ represents the vector result for $F$ and $h$ represents the vector result for $H$.

Then the adversarial attack transferability usually can be expressed as follows:
\begin{equation}\label{Eq:Transferattack}
  F^1(x) = F^2(x) = y \wedge F^1(\hat{x}) \neq y \wedge F^2(\hat{x}) \neq y
\end{equation}
This expression can be understood that for an adversarial example $\hat{x}$ which can successfully attack network $F^1$, $\hat{x}$ can also successfully attack network $F^2$.

Then perturbation $\delta = \hat{x} - x$ is generated by the adversarial example generation method which is marked as $\mathcal{A} (\cdot)$. 
Under iteration scope, the perturbation $\delta$ can be decomposed with the sum of the perturbation from each iteration which can be presented as $\delta = \sum_{t=1}^{T} \delta_t$.
The constraint for each iteration perturbation $\delta_t$ can be presented as ${||\delta_t||}_p \leq \epsilon_t$. In the I-FGSM-based method, $\epsilon_t$ usually constantly equals a certain value, e.g.  $\epsilon_t = \epsilon / T$. To make a more clear statement, with no special instructions, $\delta$ with subscript numbers represents the corresponding iteration perturbation, e.g. $\delta_t$ is the $t$-th round perturbation. And $\delta$ with superscript numbers represents the sum of the corresponding iteration perturbation, e.g. $\delta^t = \sum_{i = 1}^{t}\delta_i$.

\subsection{Looking from the future (new)}
The terminal goal of the adversarial attack is to find an optimal adversarial example that can successfully attack the target model with minor perturbation as much as it can. This task is usually transferred into an optimization task with setting a loss function as the optimization goal, e.g. the optimization goal for the untargeted attack task can be formulated as:
\begin{equation}\label{opt: untargeted}
        \underset{\delta}{\operatorname{arg max}} \quad  \mathcal{L} (F(x+\delta),y_t), s.t. ||\delta||_p \leq \epsilon,
\end{equation}
where $\epsilon$ is the threshold for the perturbation which emphasizes the perturbation constraint. Previous methods try to enhance the adversarial transferability by optimizing the optimal process with various approaches. However, from the goal to look at the optimization process, if there is an optimal point $\delta^\mathcal{O}$ for (\ref{opt: untargeted}), then the shortest search routine can be determined, e.g. the straight line segment from $0$ and $\delta^\mathcal{O}$ in Euclidean space. Then the best first iteration result $\delta^\mathfrak{B}_1$ for the entire optimization task should satisfy: 
\begin{equation}
        \underset{\delta^\mathfrak{B}_1}{\operatorname{arg min}} \quad ||\delta^\mathcal{O} - \delta^\mathfrak{B}_1||_p, s.t. ||\delta^\mathfrak{B}_1||_p \leq \epsilon_1.
\end{equation}
The solution for this optimization is
\begin{equation}
        \delta^\mathfrak{B}_1 = \alpha \cdot \delta^\mathcal{O},
\end{equation}
where $\alpha$ is a linear coefficient to limit $\delta^\mathfrak{B}_1$ satisfying the constraint. However, directly obtaining $\delta^\mathcal{O}$ is a really tough task. Finding the optimization point through iteration is one of the most effective and commonly used methods. These methods continuously approach the optimization target in an iterative manner to obtain an approximation of the optimization result. After a certain round of search, the optimization result can be treated as an approximation point $\delta^\mathcal{O'}$ for the optimal point $\delta^\mathcal{O}$. Then, the shortest search routine from $0$ to $\delta^\mathcal{O'}$ can be determined. When back to the very first iteration, the best first optimization search result should be
\begin{equation}
	\underset{\delta^\mathfrak{B'}_1}{\operatorname{arg min}} \quad ||\delta^\mathcal{O} - \delta^\mathfrak{B'}_1||_p, s.t. ||\delta^\mathfrak{B'}_1||_p = \epsilon_1.
\end{equation}
The corresponding solution for this optimization is 
\begin{equation}\label{Eq:approximation}
	\delta^\mathfrak{B'}_1 = \alpha \cdot \delta^\mathcal{O'} = \alpha \cdot {\sum_{T}^{t=1}\delta^\mathcal{O'}_t}.
\end{equation}
From a general scope, for any optimization task, after obtaining the original search routine, the approximation for the best first iteration result can also be obtained with Eq. (\ref{Eq:approximation}). Then with the same optimization process, the approximation for the best next iteration result ($\delta^\mathfrak{B'}_2$) as well as any $t$-th iteration result ($\delta^\mathfrak{B'}_t$) can be obtained. Considering that the $\delta^\mathcal{O}$ usually cannot be directly obtained, $\delta^\mathfrak{B}_t$ can also be represented with $\delta^\mathfrak{B'}_t$ in this manuscript. We name this process as Looking From the Future (LFF). 

\subsection{LFF in transfer attack}

I-FGSM-based attack methods obtain great success in transfer attacks. Intuitively, LFF can be directly used to enhance these kinds of attack methods. However, FGSM-based methods have the mechanism of symbolization of gradients which means perturbations of each iteration all satisfy $||\delta_t||_{\infty} = \epsilon_{\infty}$ and $||\delta_t||_2 = \sqrt{N \cdot \epsilon^2_{\infty}}$ at the same time. However, direcly applying Eq. (\ref{Eq:approximation}) which refers to $\delta^\mathfrak{B}_1 = {\sum_{T}^{t=1}\delta^t} / T$ will cause $||\delta^\mathfrak{B}_t||_{\infty} \leq \epsilon_{\infty}$ and $||\delta^\mathfrak{B}_1||_2 \leq ||\delta_t||_2$. Only when $\forall  i, j \in T, \delta_i = \delta_j$, $||\delta^\mathfrak{B}_1||_{\infty} = ||\delta_t||_{\infty} $ and $||\delta^\mathfrak{B}_1||_2 = ||\delta_t||_2$. 
This can only happen on a strictly linear function, while feature space in deep learning networks is usually nonlinear. 
The $\alpha$ in Eq. (\ref{Eq:approximation}) can apply $\delta^\mathfrak{B}_1$ to satisfy $||\delta^\mathfrak{B}_1||_{\infty} = ||\delta_t||_{\infty} $ or $||\delta^\mathfrak{B}_1||_2 = ||\delta_t||_2$.
However, simply applying LFF on $L_{\infty}$ constraint will not work. Keeping $||\delta^\mathfrak{B}_1||_{\infty} = ||\delta_t||_{\infty} $ will cause $||\delta^\mathfrak{B}_1||_2 \leq ||\delta_t||_2$ which means $\delta^\mathfrak{B}_t$ will have a smaller Euclidean distance compared with $\delta_t$.
A smaller Euclidean distance will lead to a slower convergence speed, which means it may even need more iterations to reach the optimal points.

Meanwhile, applying LFF on $L_2$ will cause the $||\delta^\mathfrak{B}_t||_{\infty} \geq \epsilon_{\infty}$. When the search routine comes near $T \cdot \epsilon_{\infty}$ (the $L_{\infty}$ constraint to the final perturbation), a large number of effective updates for those dimensions that already reach $T \cdot \epsilon_{\infty}$ will be discarded. This will also make the optimization routine less effective.

Also, considering that greater adversarial transferability related to more generalized examples for each model, when $\mathcal{A} (\cdot)$ itself gets overfitting to a certain model or repeatedly using history information, LFF will further exacerbate the degree of the overfitting.

\section{Methodology}
\subsection{One Order LFF Attack}
Giving an I-FGSM based attacking method $\mathcal{A}(\cdot)$ and the clean data $x$, the original $t$-th iteration from the attack process is $\delta_t = \mathcal{A} (F, x+\sum_{i=0}^{t-1}\delta_{i})$, where $t \in T$ and $x + \delta_{0} = x$. 
The gradient corresponding to the $\delta_t$ is $g_t = \mathcal{A}_g(F, x + \sum_{i=0}^{t-1}\delta_{i})$, which means that $\delta_t = \textit{sign}(g_t)$, where $\textit{sign}(\cdot)$ is the symbolization function. 
$\mathcal{Q} $ is the quantity of the steps looking from the future, i.e. $\mathcal{A}(\cdot)$ iters $\mathcal{Q}$ steps. Then the straightforward description for the $t$-th perturbation from one order LFF attack is : 
\begin{equation}\label{eq:OO}
        \delta^\mathfrak{B}_t = \alpha \cdot \textit{sign}(\sum_{q = 1}^{\mathcal{Q}} \beta ^ q \cdot \frac{\mathcal{A}_g(F, x+\sum_{j=0}^{t-1} \delta^\mathfrak{B}_{j}+\sum_{i=0}^{q-1}\delta_i)}{{||\mathcal{A}_g(F, x+\sum_{j=0}^{t-1} \delta^\mathfrak{B}_{j}+\sum_{i=0}^{q-1}\delta_i)||_p}}),
\end{equation}
where $\alpha$ is the updating rate, $\delta^\mathfrak{B}_{0} = \delta_{0}$, and $\beta$ is the future penalty coefficient.

When Eq. (\ref{eq:OO}) is directly applied to the pure I-FGSM method, $g_t$ equals the gradients of the corresponding input. The description is:
\begin{equation}
	\mathcal{A}_g(F, x + \sum_{i=0}^{t-1}\delta_{i}) = \nabla_{x + \sum_{i=0}^{t-1}\delta_{i}} \mathcal{L},
\end{equation}
where $\nabla_{x} \mathcal{L}$ is the gradient of $x$ for the loss function $\mathcal{L}$.
Then Eq. (\ref{eq:OO}) can be rewritten with the following description:
\begin{equation}
	\delta^\mathfrak{B}_t = \alpha \cdot \textit{sign}( \sum_{q = 1}^{\mathcal{Q}} \beta ^ q \cdot \frac{\nabla_{x+\sum_{j=0}^{t-1} \delta^\mathfrak{B}_{j}+\sum_{i=0}^{q-1}\delta_i} \mathcal{L}}{{||\nabla_{x+\sum_{j=0}^{t-1} \delta^\mathfrak{B}_{j}+\sum_{i=0}^{q-1}\delta_i} \mathcal{L}||_p}}),
\end{equation}

When Eq. (\ref{eq:OO}) is applied to the MI-FGSM method, which is the most widely used method, $g_t$ equals the momentums of the corresponding input. The description is
\begin{equation}\label{eq:momentum}
	\begin{aligned}
		\mathcal{A}_g(F, x + \sum_{i=0}^{t-1}\delta_{i}) & = M({x + \sum_{i=0}^{t-1}\delta_{i}}),\\
		M({x + \sum_{i=0}^{t-1}\delta_{i}}) &= \mu M({x + \sum_{i=0}^{t-2}\delta_{i}}) + \frac{\nabla_{x + \sum_{i=0}^{t-1}\delta_{i}} \mathcal{L}}{||\nabla_{x + \sum_{i=0}^{t-1}\delta_{i}} \mathcal{L}||_1},
	\end{aligned}
\end{equation}
where $M(x + \delta_0) = \frac{\nabla_x \mathcal{L}}{||\nabla_x \mathcal{L}||_1}$ and $\mu$ is the momentum decay factor. When Eq.(\ref{eq:momentum}) is introduced into Eq.(\ref{eq:OO}), the formulation can be obtained as 
\begin{equation}\label{eq:LLF_mom}
	\begin{aligned}
		\delta^\mathfrak{B}_t &= \alpha \cdot \textit{sign}(\sum_{q = 1}^{\mathcal{Q}} \beta ^ q \cdot \frac{M(x+\sum_{j=0}^{t-1} \delta^\mathfrak{B}_{j}+\sum_{i=0}^{q-1}\delta_i)}{{||M(x+\sum_{j=0}^{t-1} \delta^\mathfrak{B}_{j}+\sum_{i=0}^{q-1}\delta_i)||_p}}) \\
	\end{aligned}
\end{equation}
The expansion of Eq.(\ref{eq:LLF_mom}) (shorten $x+\sum_{j=0}^{t-1} \delta^\mathfrak{B}_{j}+\sum_{i=0}^{q-1}\delta_i$ with $x_{\mathfrak{B}_{t-1}}^{q-1}$) is 
\begin{equation}\label{eq:LLF_mom_exp}
	\begin{aligned}
		\delta^\mathfrak{B}_t &= \alpha \cdot \textit{sign}(\sum_{q = 1}^{\mathcal{Q}} \beta ^ q \cdot \frac{M(x_{\mathfrak{B}_{t-1}}^{q-1})}{{||M(x_{\mathfrak{B}_{t-1}}^{q-1})||_p}}) \\
		& = \alpha \cdot \textit{sign}(\sum_{q = 1}^{\mathcal{Q}} \frac{\beta ^ q}{{||M(x_{\mathfrak{B}_{t-1}}^{q-1})||_p}} \cdot \sum_{i=0}^{q-1} \mu ^ {q-1-i} \frac{\nabla_{x_{\mathfrak{B}_{t-1}}^{i}} \mathcal{L}}{||\nabla_{x_{\mathfrak{B}_{t-1}}^{i}} \mathcal{L}||_1}) \\
		& =  \alpha \cdot \textit{sign}( \sum_{i = 0}^{\mathcal{Q}-1} 
		(\sum_{l=i}^{\mathcal{Q}-1} \frac{\beta^{l+1} \cdot \mu^{l-i}}{||M(x_{\mathfrak{B}_{t-1}}^{l})||_p})
		 \cdot \frac{\nabla_{x_{\mathfrak{B}_{t-1}}^{i}} \mathcal{L}}{||\nabla_{x_{\mathfrak{B}_{t-1}}^{i}} \mathcal{L}||_1}).
	\end{aligned}
\end{equation}

Eq.(\ref{eq:LLF_mom_exp}) indicates that for a given optimization process, the $\delta^\mathfrak{B}_t$ can be represented as a polynomial refers to the gradient of each optimization point using $L_1$-norm regularization. The coefficient refers to the $i$-th optimization point can be marked as $\mathcal{C}_i = \sum_{l=i}^{\mathcal{Q}-1} \frac{\beta^{l+1} \cdot \mu^{l-i}}{||M(x_{\mathfrak{B}_{t-1}}^{l})||_p}$ and the gradient of $i$-th optimization point with $L_1$-norm regularization can be marked as $\mathcal{G}_i = \frac{\nabla_{x_{\mathfrak{B}_{t-1}}^{i}} \mathcal{L}}{||\nabla_{x_{\mathfrak{B}_{t-1}}^{i}} \mathcal{L}||_1}$.
Then Eq.(\ref{eq:LLF_mom_exp}) can be refined as 
\begin{equation}\label{eq:LLF_mom_vec}
	\begin{aligned}
		\delta^\mathfrak{B}_t = \alpha \cdot \textit{sign}( \left\langle \vec{\mathcal{C}}, \vec{\mathcal{G}}\right\rangle ),
	\end{aligned}
\end{equation}
where $\left\langle\cdot , \cdot\right\rangle $ is the inner product, $\vec{\mathcal{C}} = (\mathcal{C}_0, \mathcal{C}_1, \cdots ,\mathcal{C}_{\mathcal{Q}-1})$  and $\vec{\mathcal{G}} = (\mathcal{G}_0, \mathcal{G}_1, \cdots ,\mathcal{G}_{\mathcal{Q}-1})$.
Then Eq.(\ref{eq:LLF_mom_vec}) can be understood as a superposition of $\mathcal{G}_i$. 
$\mathcal{G}_i$ with a smaller value of $i$ is closer to the starting point $x$. The closer the gradient of a point is to $x$, the more local optimization information it contains; the farther the gradient of a point is from $x$, the more generalization information it contains. When coefficients of those $\mathcal{G}_i$ with a smaller value of $i$ are too great, $\delta^\mathfrak{B}_t$ will tend to $x_{\mathfrak{B}_{t-1}}^{1}$ which means overfitting. When coefficients of those $\mathcal{G}_i$ with a greater value of $i$ are too great, $\delta^\mathfrak{B}_t$ will reduce attack capability due to too much generalization. Looking back to $\vec{\mathcal{C}}$, the result of dividing $\mathcal{C}_0$ by $\mathcal{C}_{\mathcal{Q}-1}$ is 

\begin{equation}\label{eq:sim}
	\begin{aligned}
		\frac{\mathcal{C}_0}{\mathcal{C}_{\mathcal{Q}-1}} &= \frac{ \sum_{l=0}^{\mathcal{Q}-1} \frac{\beta^{l+1} \cdot \mu^{l-0}}{||M(x_{\mathfrak{B}_{t-1}}^{l})||_p} }{ \sum_{l=\mathcal{Q}-1}^{\mathcal{Q}-1} \frac{\beta^{l+1} \cdot \mu^{l-(\mathcal{Q}-1)}}{||M(x_{\mathfrak{B}_{t-1}}^{l})||_p} } \\
		&= \frac{||M(x_{\mathfrak{B}_{t-1}}^{\mathcal{Q}-1})||_p}{\beta^{\mathcal{Q}}} 
		\sum_{l=0}^{\mathcal{Q}-1} 
		\frac{\beta^{l+1} \cdot \mu^{l}}{||M(x_{\mathfrak{B}_{t-1}}^{l})||_p} \\
		&= \sum_{l=0}^{\mathcal{Q}-1} \frac{\mu^{l} \cdot \frac{||M(x_{\mathfrak{B}_{t-1}}^{\mathcal{Q}-1})||_p}{||M(x_{\mathfrak{B}_{t-1}}^{l})||_p}}{\beta^{\mathcal{Q} - l - 1}}
	\end{aligned}
\end{equation}

For an attack method without overfitting, it can be assumed that $\forall i,j \in \mathcal{Q}-1, i < j, ||M(x_{\mathfrak{B}_{t-1}}^{i})||_p < ||M(x_{\mathfrak{B}_{t-1}}^{j})||_p$. Then for given $\mu > 0$ and $\beta > 0$, $\frac{\mathcal{G}_0}{\mathcal{G}_{\mathcal{Q}-1}}$ will be greater when $\mathcal{Q}$ becomes greater. For example, when $\mu=1$ and $\beta=1$, the upper bound of Eq.(\ref{eq:sim}) is 
\begin{equation}
	\frac{\mathcal{C}_0}{\mathcal{C}_{\mathcal{Q}-1}} = \sum_{l=0}^{\mathcal{Q}-1} \frac{||M(x_{\mathfrak{B}_{t-1}}^{\mathcal{Q}-1})||_p}{||M(x_{\mathfrak{B}_{t-1}}^{l})||_p} \leq \sum_{l=0}^{\mathcal{Q}-1} \frac{\mathcal{Q}}{l+1}.
\end{equation}
Even if  $\mathcal{Q}$ is not great, e.g. $\mathcal{Q} = 5$, $\frac{\mathcal{G}_0}{\mathcal{G}_{\mathcal{Q}-1}}$ will be $11.417$ which is much greater than $1$.
And, this will cause the LFF result to tend to the $\mathcal{G}_i$ with a smaller index value. In extreme cases, this situation can be described as:
\begin{equation}
	\mathcal{Q} \to +\infty, \delta^\mathfrak{B}_t \to \alpha \cdot \textit{sign}( \mathcal{G}_0 ).
\end{equation}
Then, LLF will degenerate to the original attack method. To avoid this situation, a very simple but effective method can be applied, which is directly applying future penalty coefficient to $\vec{\mathcal{G}}$. Then the $i$-th coefficient equals $\beta ^ i$. To avoid ambiguity in the expression, this new coefficient is marked as ${\mathcal{C}'}_i = \beta ^ i$. Then the coefficients vector is $\vec{\mathcal{C}'} = (\beta^ 1, \beta^2, \dots, \beta^\mathcal{Q})$.
The perturbation update process can be described as 

\begin{equation}\label{eq:def_LFF}
	\begin{aligned}
		\delta^\mathfrak{B}_t = \alpha \cdot \textit{sign}( \left\langle \vec{\mathcal{C}'}, \vec{\mathcal{G}}\right\rangle ).
	\end{aligned}
\end{equation}

At the same time, $\delta = \sum_{i=0}^{\mathcal{Q}}\delta_{i}$ is just the approximation to the optimal point $\delta^\mathcal{O}$ with $\mathcal{Q}$ steps.  Greater $\mathcal{Q}$ refers to greater complexity. Therefore, LFF can set $\mathcal{Q}$ with a certain value that is not too great, which can combine the advantages of LFF and the low calculation complexity with the original iteration process. Especially, when $\mathcal{Q} = 1$, the LFF attack will be degenerated to the corresponding IFGSM-based method.
Naturally, the iteration between each $\delta^\mathfrak{B}_t$ can also be optimized with a similar gradient-based method, e.g. the momentum method. Then a momentum version of the LFF attack, which is named MLFF, can be described as:

\begin{equation}\label{eq:MLFF}
	\begin{aligned}
			&g^\mathfrak{B}_t = \left\langle \vec{\mathcal{C}'}, \vec{\mathcal{G}}\right\rangle,\\
			&M^\mathfrak{B}_t = \eta \cdot M^\mathfrak{B}_{t-1} + \frac{g^\mathfrak{B}_t}{{||g^\mathfrak{B}_t||}_1},\\
			&\delta^\mathfrak{B}_t = \alpha \cdot \textit{sign}(M^\mathfrak{B}_t),
	\end{aligned}
\end{equation}
where $\eta$ is the momentum decay factor.

\subsection{Multi-order LFF ($\textit{LFF}^\mathcal{N}$) Attack}

The sequence of the $\delta^\mathfrak{B}_t$ from the one-order LFF attack can still be enhanced with the LFF mechanism. With the same guidelines, the enhancing process can be iterated to $\mathcal{N}$-order LFF, which is named $\textit{LFF}^\mathcal{N}$ attack. The description for the $\textit{LFF}^\mathcal{N}$ attack is:
\begin{equation}\label{eq:MO}
        \delta^{\mathfrak{B}^\mathcal{\nu} }_t = \alpha \cdot \textit{sign}( \left\langle \vec{{\mathcal{C}^\mathcal{\nu}}'}, \vec{\mathcal{G}^\mathcal{\nu}}\right\rangle ),
\end{equation}
where $\nu \in \mathcal{N}$ is the indicator for the order,
\begin{equation}
	\begin{aligned}
		{\mathcal{C}^\mathcal{\nu}}'_i &= {\beta^\mathcal{\nu}}^i, \\
		\mathcal{G}^\mathcal{\nu}_t &= 
		\frac{\nabla_{x+\sum_{j=0}^{t-1} \delta^{\mathfrak{B}^\nu}_{j}+\sum_{i=0}^{q-1}\delta^{\mathfrak{B}^{\nu-1}}_i} \mathcal{L}}
		{||\nabla_{x+\sum_{j=0}^{t-1} \delta^{\mathfrak{B}^\nu}_{j}+\sum_{i=0}^{q-1}\delta^{\mathfrak{B}^{\nu-1}}_i} \mathcal{L}||_1},
	\end{aligned}
\end{equation}
and $\delta^{\mathfrak{B}^{0}}_i = \delta_i$.
Especially, when $\mathcal{N} = 1$, Eq.(\ref{eq:MO}) will be degenerated to Eq.(\ref{eq:def_LFF}). In the $\textit{LFF}^\mathcal{N}$ attack, there are multi-iteration processes. Therefore, there are many ways to combine operations that can further optimize the optimization process. Here, only a simple extension for MLFF with applying momentum mechanism for any $\nu$-th order, where $\nu \in \mathcal{N}$. This method is named $\textit{MLFF}^\mathcal{N}$. The corresponding formulation is shown as follows:

\begin{equation}\label{eq:MLFFN}
        \begin{aligned}
                &g^{\mathfrak{B}^\mathcal{\nu} }_t = \left\langle \vec{{\mathcal{C}^\mathcal{\nu}}'}, \vec{\mathcal{G}^\mathcal{\nu}}\right\rangle,\\
                &M^{\mathfrak{B}^\mathcal{\nu}}_t = \eta_{\mathcal{\nu}} \cdot M^{\mathfrak{B}^\mathcal{\nu}}_{t-1} + \frac{g^{\mathfrak{B}^\mathcal{\nu} }_t}{{||g^{\mathfrak{B}^\mathcal{\nu} }_t||}_1},\\
                &\delta^{\mathfrak{B}^\mathcal{\nu} }_t = \alpha \cdot \textit{sign}(M^{\mathfrak{B}^\mathcal{\nu}}_t),
        \end{aligned}
\end{equation}
where $\eta_{\mathcal{\nu}}$ is the momentum penalty factor for $\nu$-th order.

\section{Experiment}
\subsection{Experiment Settings}
Because most SOTA methods are realized on MI-FGSM, we conduct the comparison experiment using the MLLF method with those SOTA methods.

\subsubsection{Classifcation Networks}
We conduct a wide range of classification networks within the Timm which is one of the greatest deep neural network libraries with various pre-trained models. We apply both CNN structure networks and Transformer structure networks. All classification networks include ResNet-50, BiT-50, Inception-v3, Inception-ResNet-v2, ConvNeXT-B, ViT-B, Swin-B, Deit-B. ResNet is one of the most classic deep learning networks. BiT-50 is one of the latest structures for ResNet models, which represents the best performance of the ResNet series. Inception-v3 is the pure inception structure CNNs, and Inception-ResNet is the combination of ResNet and inception structure. ConvNeXT is the most latest CNN structure network which represents the best performance of CNNs. Vision Transformer(ViT) is the typical Transformer structure network for image classification. Swin and Deit are two of the latest Transformer structure models. 

\subsubsection{Dataset}
We conduct the ImageNet1k dataset for all experiments. ImageNet1k dataset contains 1000 classes from the ImageNet dataset. All classification networks used in the experiments from Timm are pre-trained with the ImageNet1k dataset. Specifically, we conduct the ILSVRC2012 dataset as the testing dataset. The ILSVRC2012 dataset is one of the most classic datasets for adversarial attacks. The ILSVRC2012 dataset comes from the testing data of the ImageNet1k dataset. It still is widely used for evaluating adversarial attack performance, especially the transfer attack.

\subsubsection{Baselines}
We conduct four attack methods as the baselines for the comparison experiment. They are MI-FGSM, EMI-FGSM, Admix and SIA. MI-FGSM is one of the most classic methods as well as one of the most widely used methods. The other baselines are all realized based on MI-FGSM. EMI-FGSM is one of the most recent gradient-based methods. Admix can typically represent the method with the mixup strategy. SIA is one of the latest self-transformation data augmentation methods. SIA is also one of the methods to achieve the best attack performance.

\begin{table*}
	\centering
	\begin{tabular}[c]{|c|c|rrrrrrrr|}
		\hline
		Model & Attack & Res50 & BiT50 & IncRes-v2 & ConvneXT-B & ViT-B & Swin-B & Deit-B & Inc-v3\\
		\hline
		\hline
		\multirow{12}*{Inc-v3} 
		  & MI 		& 41.4&	20.6&	43.9&	20.6&	9.3&	15.1&	11.3&	100.0  \\
		~ & MLFF-MI & 47.0&	23.4&	55.5&	24.4&	9.7&	16.2&	12.5&	100.0  \\
		~ & Inc.	& \textbf{+5.6}&	\textbf{+2.8}&  \textbf{+11.6}&	\textbf{+3.8}&  \textbf{+0.4}&	\textbf{+1.1}&	\textbf{+1.2}&	+0.0  \\
		\cline{2-10}
		~ & EMI     & 71.1&	43.6&	81.0&	42.0&	16.9&	33.0&	22.4&	100.0 \\
		~ & MLFF-EMI& 76.2&	49.4&	85.1&	48.0&	21.9&	37.8&	26.6&	100.0 \\
		~ & Inc.	& \textbf{+5.1}&	\textbf{+5.8}&	\textbf{+4.1}&	\textbf{+6.0}&	\textbf{+5.0}&	\textbf{+4.8}&	\textbf{+4.2}&	+0.0  \\
		\cline{2-10}
		~ & Admix      & 80.0&	52.6&	86.8&	45.9&	23.3&	39.3&	30.5&	100.0 \\	
		~ & MLFF-Admix & 82.7&	53.6&	87.5&	46.2&	23.6&	39.6&	30.1&	100.0 \\
		~ & Inc.	   & \textbf{+2.7}&	\textbf{+1.0}&	\textbf{+0.7}&	\textbf{+0.3}&	\textbf{+0.3}&	\textbf{+0.3}&	-0.4 &	+0.0  \\
		\cline{2-10}	
		~ & SIA      & 91.9 &	72.3&	95.8&	62.6&	32.2&	54.5&	36.0&	100.0 \\
		~ & MLFF-SIA & 95.9	&	81.6&	99.0&	74.0&	40.0&	63.9&	43.2&	100.0\\
		~ & Inc.	 &\textbf{+4.0}&\textbf{+9.3}&\textbf{+3.2}&\textbf{+11.4}&\textbf{+7.8}&\textbf{+9.4}&\textbf{+6.8}& +0.0  \\
		\hline
		\multirow{12}*{ViT-B} 
		  & MI		& 51.3&	37.7&	32.6&	41.2&	100.0&	54.4&	66.9&	39.8 \\
		~ & MLFF-MI & 53.1&	37.4&	32.9&	38.3&	100.0&	52.0&	68.4&	42.4 \\
		~ & Inc.	&\textbf{+1.8}&\textbf{-0.3}&\textbf{+0.3}& -2.9 & +0.0 & -2.2 &\textbf{+1.5}& \textbf{+2.6}  \\
		\cline{2-10}
		~ & EMI 	 & 71.8&	62.2&	55.8&	66.7&	100.0&	77.8&	89.9&	62.5\\		
		~ & MLFF-EMI & 76.2&	69.7&	60.4&	70.2&	100.0&	82.5&	94.2&	67.0\\
		~ & Inc.	 &\textbf{+4.4}&\textbf{+7.5}&\textbf{+4.6}&\textbf{+3.5}& +0.0 & \textbf{+4.7} &\textbf{+4.3}& \textbf{+4.5}  \\
		\cline{2-10}
		~ & Admix 		& 68.1&	61.2&	51.6&	63.6&	99.9&	76.0&	87.4&	58.5 \\		
		~ & MLFF-Admix 	& 72.9&	64.2&	53.7&	64.7&	100.0&	79.4&	91.2&	61.8 \\
		~ & Inc.	 	&\textbf{+4.8}&\textbf{+3.0}&\textbf{+2.1}&\textbf{+1.1}&\textbf{+0.1}&\textbf{+3.4}&\textbf{+3.8}& \textbf{+3.3}  \\
		\cline{2-10}
		~ & SIA 		& 82.6&	80.7&	71.2&	82.5&	99.6&	87.5&	89.0&	75.6 \\		
		~ & MLFF-SIA 	& 91.1&	90.8&	83.8&	91.9&	100.0	&94.5&	96.2&	86.9 \\
		~ & Inc.	 	&\textbf{+8.5}&\textbf{+10.1}&\textbf{+12.6}&\textbf{+9.4}&\textbf{+0.4}&\textbf{+7.0}&\textbf{+7.2}&\textbf{+11.3}  \\
			
		 \hline
	\end{tabular}
	\\ [0.2cm]
	\large
	\captionsetup{justification=justified}
	\caption{The attack success rates (\%) against eight networks by baseline attacks and our method. The Increment (Inc. \%) to the corresponding baselines that are greater than zero are marked in bold.}
	\label{untarget_single_surrogate_model}
\end{table*}

\begin{table*}
	\centering
	\small
	\begin{tabular}[c]{|c|c|cccccc|}
		\hline
		Model & Attack & Inc-v3$_{adv}$ & IncRes-v2$_{adv}$ & ConvNeXT-B+FD & Swin-B+FD & ConvNeXT-B+Bit-Red & Swin-B+Bit-Red\\
		\hline
		\hline
		\multirow{12}*{Inc-v3}
		  & MI 		& 19.7&	7.8&	16.5&	15.2&	20.8&	14.7\\
		~ & MLFF-MI & 23.4&	8.8&	16.7&	16.1&	23.7&	16.4\\
		~ & Inc. 	& \textbf{+3.7}&	\textbf{+1.0}&	\textbf{+0.2}&	\textbf{+0.9}&	\textbf{+2.9}&	\textbf{+1.7}\\
		\cline{2-8}
		~ & EMI 	 & 38.5&	15.6&	27.8&	25.4&	42.2&	33.0\\
		~ & MLFF-EMI & 43.3&	19.4&	31.0&	28.8&	46.7&	37.9\\
		~ & Inc. 	& \textbf{+4.8}&	\textbf{+3.8}&	\textbf{+3.2}&	\textbf{+3.4}&	\textbf{+4.5}&	\textbf{+4.9}\\
		\cline{2-8}
		~ & Admix 		& 55.2&	27.2&	36.9&	33.0&	45.6&	39.5\\	
		~ & MLFF-Admix  & 55.8&	26.9&	37.2&	34.0&	46.6&	40.9\\
		~ & Inc. 	& \textbf{+0.6}&	\textbf{-0.3}&	\textbf{+0.3}&	\textbf{+1.0}&	\textbf{+1.0}&	\textbf{+1.4}\\
		\cline{2-8}
		~ & SIA 		& 54.0&	28.2&	39.3&	38.6&	61.8&	53.2\\
		~ & MLFF-SIA    & 68.6&	35.1&	42.3&	42.5&	73.7&	65.1\\ 
		~ & Inc. 	& \textbf{+14.6}&	\textbf{+6.9}&	\textbf{+3.0}&	\textbf{+3.9}&	\textbf{+11.9}&	\textbf{+11.9}\\
		\hline
	\end{tabular}
	\\ [0.2cm]
	\large
	\captionsetup{justification=justified}
	\caption{The attack success rates (\%) against six defensive models by baseline attacks and our method. The Increment (Inc. \%) to the corresponding baselines that are greater than zero are marked in bold.}
	\label{untargeted_defensive_models}
\end{table*}

\subsubsection{Hyperparameters}
Following many previous works, the perturbation constraint is set to $L_{\infty} = 16$, the quantity of the iteration is $16$, the update rate is $\alpha = \frac{1}{255}$. 
For the very basic method MI-FGSM, the momentum decay factor is $1$. The number of gradient collections in EMI-FGSM is $11$ with a linear sample and the radius value is $7$. For Admix, the number of scale copies is set to $5$ and the number of mix images is set to $3$ while the mix ratio is $0.2$. In SIA, we set the splitting number to $3$ and the number of transformed images for gradient calculation is $10$.

In the comparison experiment, to fairly compare the performance of LFF and other methods, $\mathcal{Q}$ is set to $4$, which means a low calculation complexity and only looking from a very near future. Because when $\mathcal{Q}$ is small LFF will tend to I-FGSM even if the attack method applied is the MI-FGSM-based method, the MLFF method is applied to the comparison experiment. The future penalty factor $\beta$ is set to $1$ and the momentum decay for the MLFF is $1$.

\subsubsection{Evaluation metrics}
Attack Success Rate (ASR) is applied as the main evaluation metric for the performance of attack methods. ASR can present how many images have been successfully attacked, no matter for the white-box network (surrogate network) or the black-box network (victim networks). In addition, because the LFF attack is applied to the existing attack methods, the increment of the ASR, which is marked as Inc., is also considered to present the increment of the performance compared with the original attack method. The greater ASR and Inc. are, the better. 

\subsection{Comparison Experiments}
The comparison experiment is applied within four scopes. Firstly we evaluate baselines and our proposed method under the condition of a single surrogate network with the untargeted attack task. Secondly, we apply the comparison on the defensive networks as well as defense methods applied to pre-trained networks. Thirdly, the targeted attack task is applied. Fourthly, ensemble networks are applied as the surrogate network. The adversarial example from ensemble networks is tested by both original networks and defensive networks as well as defense methods.

\begin{table*}
	\centering
	\small
	\begin{tabular}[c]{|c|c|cccccccc|}
		\hline
		Model & Attack & Res50 & BiT50 & IncRes-v2 & ConvNeXT-B & ViT-B & Swin-B & Deit-B & Inc-v3\\
		\hline
		\hline
		\multirow{12}*{Inc-v3} 
		  & MI 		& 0.3&	0.4&	0.5&	0.4&	0.0&	0.1&	0.1&	100.0\\
		~ & MLFF-MI & 0.6&	0.4&	0.9&	0.2&	0.1&	0.0&	0.1&	100.0\\
		~ & Inc.    &\textbf{+0.3}&+0.0&\textbf{+0.4}&-0.2&\textbf{+0.1}&-0.1&+0.0&+0.0\\
		\cline{2-10}
		~ & EMI 		& 2.7&	1.2&	5.9&	1.4&	0.1&	0.7&	0.2&	97.4\\
		~ & MLFF-EMI 	& 5.0&	2.0&	12.1&	2.0&	0.4&	1.1&	1.0&	100.0\\
		~ & Inc.    &\textbf{+2.3}&\textbf{+0.8}&\textbf{+6.2}&\textbf{+0.6}&\textbf{+0.3}&\textbf{+0.4}&\textbf{+0.8}&\textbf{+2.6}\\
		\cline{2-10}
		~ & Admix 		& 2.9&	0.7&	5.1&	1.2&	0.4&	0.5&	0.3&	98.7\\	
		~ & MLFF-Admix  & 4.1&	1.7&	8.7&	1.7&	0.5&	0.8&	0.5&	99.8\\
		~ & Inc.    &\textbf{+1.2}&\textbf{+1.08}&\textbf{+3.6}&\textbf{+0.5}&\textbf{+0.1}&\textbf{+0.3}&\textbf{+0.2}&\textbf{+1.1}\\
		\cline{2-10}	
		~ & SIA 		& 7.0&	3.8&	15.9&	3.9&	1.0&	2.6	&1.2&	74.5\\
		~ & MLFF-SIA 	& 26.7&	13.7&	49.3&	13.2&	2.3&	6.8 &3.5&	97.0\\ 
		~ & Inc.    &\textbf{+19.7}&\textbf{+9.9}&\textbf{+33.4}&\textbf{+8.3}&\textbf{+1.3}&\textbf{+4.2}&\textbf{+2.3}&\textbf{+22.5}\\
		\hline
	\end{tabular}
	\\ [0.2cm]
	\large
	\captionsetup{justification=justified}
	\caption{The attack success rates (\%) against eight models by baseline attacks and our method for the targeted attack task. The Increment (Inc. \%) to the corresponding baselines that are greater than zero are marked in bold.}
	\label{targetted_attack}
\end{table*}

\begin{table*}
	\centering
	\small
	\begin{tabular}[c]{|c|c|cccccccc|}
		\hline
		Model & Attack & Res50 & BiT50 & IncRes-v2 & ConvNeXT-B & ViT-B & Swin-B & Deit-B & Inc-v3\\
		\hline
		\hline
		\multirow{12}*{Em} 
			& MI 		& 100.0&	68.9&	72.4&	70.4&	98.5&	71.0&	98.9&	99.7\\
		~ 	& MLFF-MI 	& 100.0&	75.4&	79.5&	73.7&	100.0&	73.8&	99.8&	100.0\\
		~	& Inc.    	&+0.0&\textbf{+6.5}&\textbf{+7.1}&\textbf{+3.3}&\textbf{+1.5}&\textbf{+2.8}&\textbf{+0.9}&\textbf{+0.3}\\
		\cline{2-10}
		~ & EMI 		& 100.0&	92.1&	94.7&	89.5&	99.6&	90.5&	99.6&	100.0\\
		~ & MLFF-EMI 	& 100.0&	95.2&	96.5&	92.8&	100.0&	93.3&	100.0&	100.0\\
		~ & Inc.    	& +0.0 &\textbf{+3.1}&\textbf{+1.8}&\textbf{+3.3}&\textbf{+0.4}&\textbf{+2.8}&\textbf{+0.4}&+0.0\\
		\cline{2-10}
		~ & Admix 		& 100.0&	93.8&	96.8&	88.7&	97.6&	90.2&	97.7&	100.0\\	
		~ & MLFF-Admix 	& 100.0&	96.2&	98.7&	93.1&	99.7&	93.8&	99.8&	100.0\\
		~ & Inc.    	&+0.0&\textbf{+2.4}&\textbf{+1.9}&\textbf{+4.4}&\textbf{+2.1}&\textbf{+3.6}&\textbf{+2.1}&+0.0\\
		\cline{2-10}	
		~ & SIA 		& 100.0&	98.1&	98.8&	97.3&	98.9&	97.3&	99.4&	99.8\\
		~ & MLFF-SIA 	& 100.0&	99.8&	99.9&	99.6&	99.8&	99.4&	99.8&	100.0\\
		~ & Inc.    	&+0.0 &\textbf{+1.7}&\textbf{+1.1}&\textbf{+2.3}&\textbf{+0.9}&\textbf{+2.1}&\textbf{+0.4}&\textbf{+0.2}\\
		\hline
	\end{tabular}
	\\ [0.2cm]
	\large
	\captionsetup{justification=justified}
	\caption{The attack success rates (\%) against eight models by baseline attacks and our method with ensembling Res50, Inc-v3, ViT-B and Deit-B. The Increment (Inc. \%) to the corresponding baselines that are greater than zero are marked in bold.}
	\label{Ensemble_models}
\end{table*}

\begin{table*}
	\centering
	\small
	\begin{tabular}[c]{|c|c|cccccc|}
		\hline
		Model & Attack & Inc-v3$_{adv}$ & IncRes-v2$_{adv}$ & ConvNeXT-B+FD & Swin-B+FD & ConvNeXT-B+Bit-Red & Swin-B+Bit-Red\\
		\hline
		\hline
		\multirow{12}*{Em} 
		  & MI 		& 37.4&	25.0&	45.6&	52.1&	69.5&	70.5\\
		~ & MLFF-MI & 40.1&	22.5&	47.3&	53.3&	73.8&	73.7\\
		~ & Inc. 	& \textbf{+2.7}&	\textbf{+2.5}&	\textbf{+1.7}&	\textbf{+1.2}&	\textbf{+4.3}&	\textbf{+3.2}\\
		\cline{2-8}
		~ & EMI 	& 69.0&	48.2&	73.3&	76.6&	89.3&	90.2\\
		~ & MLFF-EMI& 76.0&	55.3&	78.7&	80.1&	92.5&	93.2\\
		~ & Inc. 	& \textbf{+7.0}&	\textbf{+7.1}&	\textbf{+5.4}&	\textbf{+3.5}&	\textbf{+3.2}&	\textbf{+3.0}\\
		\cline{2-8}
		~ & Admix 		& 80.6&	64.0&	77.6&	79.9&	89.1&	89.8\\	
		~ & MLFF-Admix 	& 85.1&	69.0&	82.2&	84.4&	93.5&	94.1\\
		~ & Inc. 	& \textbf{4.5}&	\textbf{+5.0}&	\textbf{+4.6}&	\textbf{+4.5}&	\textbf{+4.4}&	\textbf{+4.3}\\
		\cline{2-8}	
		~ & SIA 		& 86.4&	72.3&	85.3&	86.7&	97.4&	97.3\\
		~ & MLFF-SIA 	& 94.0&	84.0&	91.5&	93.4&	99.6&	99.4\\ 
		~ & Inc. 	& \textbf{+7.6}&	\textbf{+11.7}&	\textbf{+6.2}&	\textbf{+6.7}&	\textbf{+2.2}&	\textbf{+2.1}\\
		\hline
	\end{tabular}
	\\ [0.2cm]
	\large
	\captionsetup{justification=justified}
	\caption{The attack success rates (\%) against six defensive models by baseline attacks and our method with ensembling Res50, Inc-v3, ViT-B and Deit-B. The Increment (Inc. \%) to the corresponding baselines that are greater than zero are marked in bold.}
	\label{Ensemble_models_defensive}
\end{table*}

\begin{figure*}
	\label{Fig: ablation1}
	\centering
	\includegraphics[width=17.5cm]{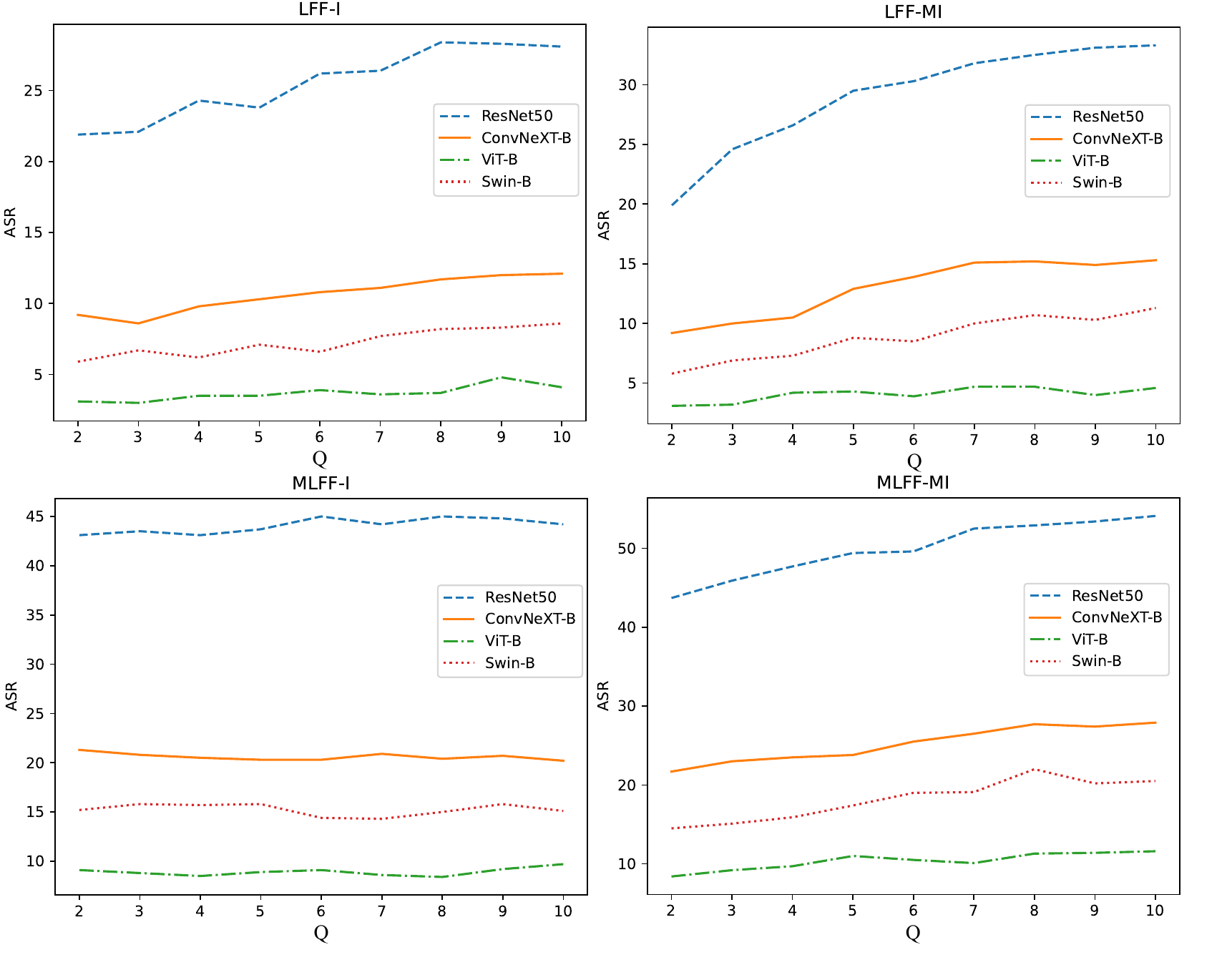}
	\caption{The experimental results of changing $\mathcal{Q}$ in LFF-I, LFF-MI, MLFF-I, and MLFF-MI respectively.}
\end{figure*}

\begin{figure*}
	\label{Fig: ablation2}
	\centering
	\includegraphics[width=17.5cm]{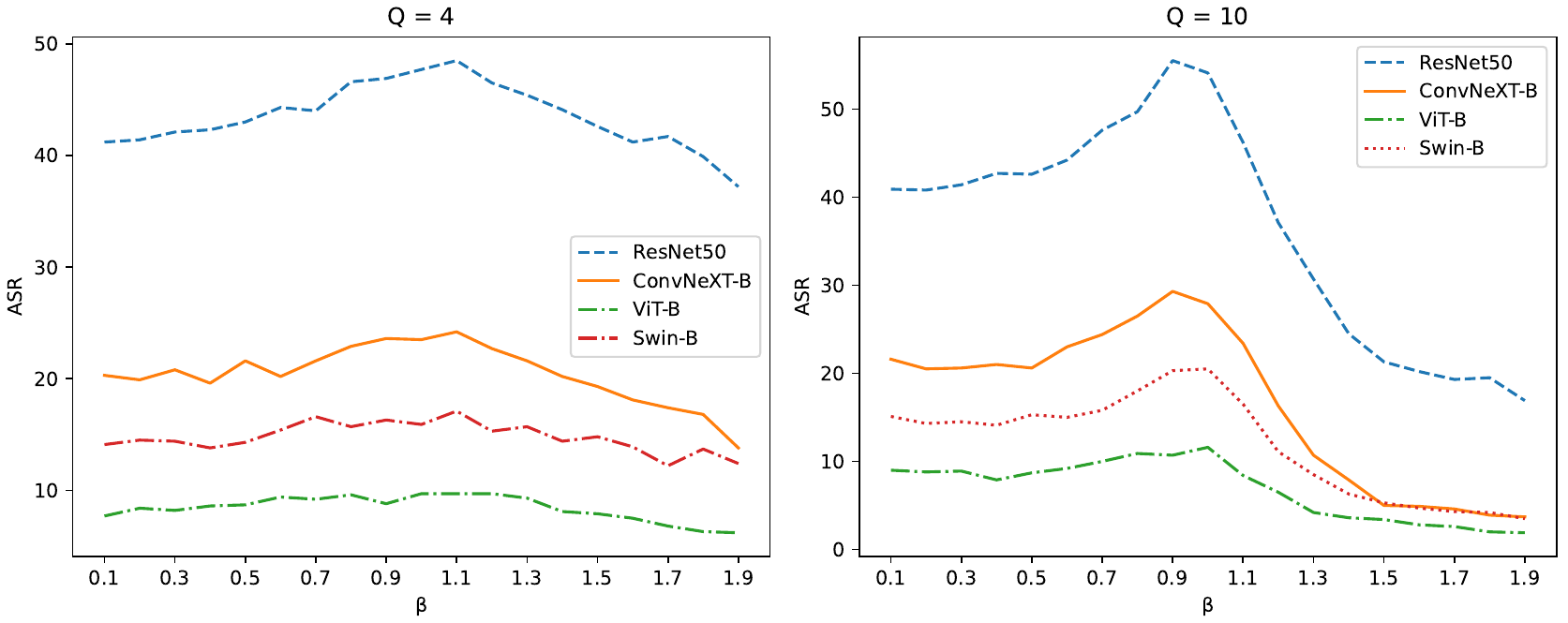}
	\caption{The experimental results of changing $\beta$ in MLFF-MI with $\mathcal{Q} = 4$ and $\mathcal{Q} = 10$ respectively.}
\end{figure*}

\subsubsection{Untargeted attack with single surrogate network}
In this comparison experiment, Inception-v3 and Vit-B are chosen to be the surrogate networks individually. Inception-v3 represents the adversarial examples from the CNN structure, while ViT represents the adversarial examples from the Transformer structure. All other networks are applied as the victim networks. The experimental results are shown in Table \ref{untarget_single_surrogate_model}.

Overall, the MLFF method can greatly improve the performance of all baselines. No matter whether the surrogate network is the CNN structure or the Transformer structure, the ASRs for the victim networks almost have increased. Especially, some of Inc. are over $10\%$. From the details, there are three conditions the Inc. is the negative value. One occurs by applying Admix as the baseline with generating adversarial examples from Inception-v3 and testing for the Deit-B networks. However, when observing the Inc. for the MLFF Admix, it can be noted that MLFF seems to have not very great performance improvement on Admix. This might refer to the randomness in Admix. For each iteration, Admix randomly chooses images to mix up. Therefore, the process of the optimization itself is not fixed. Especially when this randomness is great, the randomness of the original sequence obtained in the LFF mechanism will be very great, thereby reducing the optimization effect. The performance of the MLFF-Admix in ViT-B can also verify this phenomenon.
Another two occur by applying MI-FGSM as the baseline with generating adversarial examples from ViT-B and testing for ConvNeXT-B and Swin-B. This may be due to the small value of $\mathcal{Q}$ we set. The small $\mathcal{Q}$ will cause the MLFF to tend to the original attack, i.e. the MI-FGSM itself. At the same time, when the attack method tramps into the overfitting, MLFF will accelerate the overfitting. This all might be the reason that there are conditions that MLFF has negative Inc. values.
It should also be noted that MLFF has very great Inc. values for both EMI and SIA methods. Although SIA itself has performed relatively well on ASR, MLFF-SIA can further enhance the adversarial transferability with a larger degree of improvement.

\subsubsection{Untargeted attack with defensive models}
In this comparison experiment, two adversarial training networks are applied, i.e. Adversarial Inception-v3 and Ensemble Adversarial Inception-ResNet-v2. Two defensive methods FD and Bit-Red are applied to the ConvNext and Swin individually. The surrogate network is the Inception-v3 network. The experimental results are shown in Table \ref{untargeted_defensive_models}.

The experimental results are still corresponding to the previous experimental results. The increment to MI-FGSM and Admix method are relatively small, while the increment to EMI and SIA are very great.

\subsubsection{Targeted attack}
In this comparison experiment, the targeted label for the attack is randomly chosen without repeating. The loss function for the targeted attack is the Cross-Entropy loss. The surrogate model is the Inception-v3 network. The experimental results are shown in Table \ref{targetted_attack}.



\subsubsection{Untargeted attack with ensemble networks}
In this comparison experiment, the ensembled network is treated as the surrogate network for adversarial example generation. The ensembled network contains four networks, i.e. ResNet50, Inception-v3, ViT-B and Deit-B. This ensemble network contains both CNN structure networks and Transformer structure networks. The attack success rates for the original networks are placed in Table \ref{Ensemble_models}. The attack success rates for the defensive networks as well as defense methods are placed in Table \ref{Ensemble_models_defensive}.


\subsection{Ablation Experiment}

The ablation experiment is applied within 2 scopes. Firstly, we verify the effectiveness of the LFF as well as the MLFF with different $\mathcal{Q}$. Then, we verify the influence of the future penalty factor $\beta$. 


\subsubsection{Influence of the looking from the future steps}
In this ablation experiment, the quantity of the looking from the future $\mathcal{Q}$ is changed to verify the mechanism of the LLF. With greater $\mathcal{Q}$, LFF attack as well as MLFF attack will have better attack performance. With smaller $\mathcal{Q}$, the LFF attack will have very poor attack performance but the MLFF attack will keep a relatively stable performance. 
We apply the LFF attack and the MLFF attack to MI-FGSM and I-FGSM respectively. The experimental results are shown in the Fig. 1.

Overall, the experimental results can sufficiently verify the effect of the $\mathcal{Q}$. The greater value of $\mathcal{Q}$ is, LFF attack as well as MLFF attack performances better. This tendency is especially obvious when the baseline is the MI-FGSM. However, this tendency seems to be weak for the MLFF-I attack. We speculate that this phenomenon is caused by the I-FGSM. Because I-FGSM easily falls into the local minima, which means low adversarial transferability, MLFF can only help I-FGSM quickly approach this local optimal point, although MLFF-I has a relatively good performance compared with the MI-FGSM.

\subsubsection{Influence of the future penalty factor}
In this ablation experiment, the future penalty factor $\beta$ is changed. With greater $\beta$, the LFF attack will contain more future information to generalize the perturbation. When $\beta$ is too great, the attack performance will drop due to losing attack capability.
With smaller $\beta$, the LFF attack will contain more local attack information. When $\beta$ is too small, the attack performance will degenerate to the original attack.
Hence we apply $\beta$ from $0.1$ to $1.9$ with $0.1$ intervel under the condition of $\mathcal{Q} = 4$ and $\mathcal{Q} = 10$ respectively. Because $\mathcal{Q} = 4$ is relatively small, the MLFF method is applied. The baseline is the MI-FGSM method. The experimental results are shown in the Fig. 2.

Overall, it can be noted that the trend in both subfigures is that ASR gradually increases with the increment of $\beta$ value. After reaching the peak, ASR gradually decreases as the $\beta$ value decreases. This phenomenon verifies the guess about the effect of the $\beta$, i.e. the effect of the gradients in Eq. (\ref{eq:def_LFF}).
A closer look at the horizontal coordinates of the peaks of the two graphs reveals that the optimal value of $\beta$ is different when facing different $\mathcal{Q}$. When $\mathcal{Q}$ becomes greater, the best $\beta$ seems to become smaller.  When comparing the two subfigures horizontally, it can be found that the average performance is significantly better when the $\mathcal{Q}$ is greater than when the $\mathcal{Q}$ is smaller.




\section{Conclusion and Discussion}
In this paper, we propose a novel concept which is named Looking From the Future. We extend the LFF concept into the adversarial attack task and propose the LFF attack as well as the MLFF attack while eliminating the disadvantages of focusing the local information and overfitting. Furthermore, we extend the LFF attack to the multi-order LFF attack, which is named $\text{LFF} ^ \mathcal{N}$. Comparison experiments on four tasks as well as ablation experiments have conducted the performance of our proposed method. Experimental results clearly show that the LFF attack can greatly increase existing adversarial attack methods.


\bibliographystyle{abbrv}
\bibliography{nips}

\end{document}